\documentclass[letterpaper, 10 pt, conference]{ieeeconf}  % Comment this line out if you need a4paper

\IEEEoverridecommandlockouts                              % This command is only needed if 
\pdfoutput=1

\usepackage{graphics} % for pdf, bitmapped graphics files
\usepackage{epsfig} % for postscript graphics files
\usepackage{times} % assumes new font selection scheme installed
\usepackage{amsmath} % assumes amsmath package installed
\usepackage{amssymb}  % assumes amsmath package installed
\usepackage{dsfont}  % Add this in the preamble

\usepackage[font=small]{caption}
\usepackage[font=footnotesize]{subcaption}  % or subfig, if required by your conference
\usepackage{url}
\usepackage{hyperref}
\hypersetup{
    colorlinks=true, 
    urlcolor=cyan,
    }
\usepackage{booktabs}

\usepackage{color}
\usepackage{xcolor}

\usepackage{caption}

\renewcommand{\baselinestretch}{0.985}

\usepackage{float}

\title{\LARGE \bf
Bench-Push: Benchmarking Pushing-based Navigation and Manipulation Tasks for Mobile Robots}

\newcommand{\projectname}{Bench-Push}

\author{Ninghan Zhong$^{1}$*, Steven Caro$^{2}$*, Megnath Ramesh$^{2}$, Rishi Bhatnagar$^{3}$, \\ Avraiem Iskandar$^{2}$, and Stephen L. Smith$^{2}$% <-this % stops a space
\thanks{*denotes equal contribution.}
\thanks{$^{1}$Institute for Robotics and Intelligent Machines, Georgia Institute of Technology, Atlanta, GA 30332, USA (e-mail: \protect\url{nzhong34@gatech.edu}). This work was done while Ninghan Zhong was a student at the University of Waterloo.}
\thanks{$^{2}$Department of Electrical and Computer Engineering, University of Waterloo, Waterloo, ON N2L 3G1, Canada (e-mail: \protect\url{{steven.caro, avraiem.iskandar, m5ramesh, stephen.smith}@uwaterloo.ca})}
\thanks{$^{3}$Department of Mechanical Engineering, University of Alberta, Edmonton, AB T8G 2R3, Canada (e-mail: \protect\url{rbhatna1@ualberta.ca}). This work was done while Rishi Bhatnagar was visiting the University of Waterloo.}
\thanks{This work is supported in part by the National Research Council Canada (NRC) and in part by the Natural Sciences and Engineering Research Council of Canada (NSERC).} 
\thanks{Resources used in preparing this research were provided, in part, by the Province of Ontario, the Government of Canada through CIFAR, and companies sponsoring the Vector Institute.}
}

\begin{document}

\maketitle
\thispagestyle{empty}
\pagestyle{empty}

\begin{abstract}
\label{sec:abstract}
Mobile robots are increasingly deployed in cluttered environments with movable objects, posing challenges for traditional methods that prohibit interaction. In such settings, the mobile robot must go beyond traditional obstacle avoidance, leveraging pushing or nudging strategies to accomplish its goals.
While research in pushing-based robotics is growing, evaluations rely on ad hoc setups, limiting reproducibility and cross-comparison. 
To address this, we present \projectname, the first unified benchmark for pushing-based mobile robot navigation and manipulation tasks. \projectname~includes multiple components: 1) a comprehensive range of simulated environments that capture the fundamental challenges in pushing-based tasks, including navigating a maze with movable obstacles, autonomous ship navigation in ice-covered waters, box delivery, and area clearing, each with varying levels of complexity; 2) novel evaluation metrics to capture efficiency, interaction effort, and partial task completion; and 3) demonstrations using \projectname~to evaluate example implementations of established baselines across environments. \projectname~will be open-sourced as a Python library with a modular design. The code, documentation, and trained models can be found at {\footnotesize \url{https://github.com/IvanIZ/BenchNPIN}}.
\end{abstract}
\section{Introduction}
\label{sec:introduction}
The key to robust mobile robot deployment lies in the robot's ability to operate in complex environments. Traditional mobile robot approaches typically focus on computing collision-free paths or actions. When mobile robots are deployed in cluttered and unstructured environments, such as homes, hospitals, or disaster sites, where movable objects might block feasible routes, finding collision-free paths is often impractical. Under such settings, the ability for robots to interact with movable objects and obstacles is essential for task completion. This leads to pushing-based navigation and manipulation, where the mobile robot must interact with objects through pushing or nudging actions to progress.

\setlength{\belowcaptionskip}{-5pt}
\begin{figure}[t]
    \centering
    \includegraphics[width=0.9\linewidth]{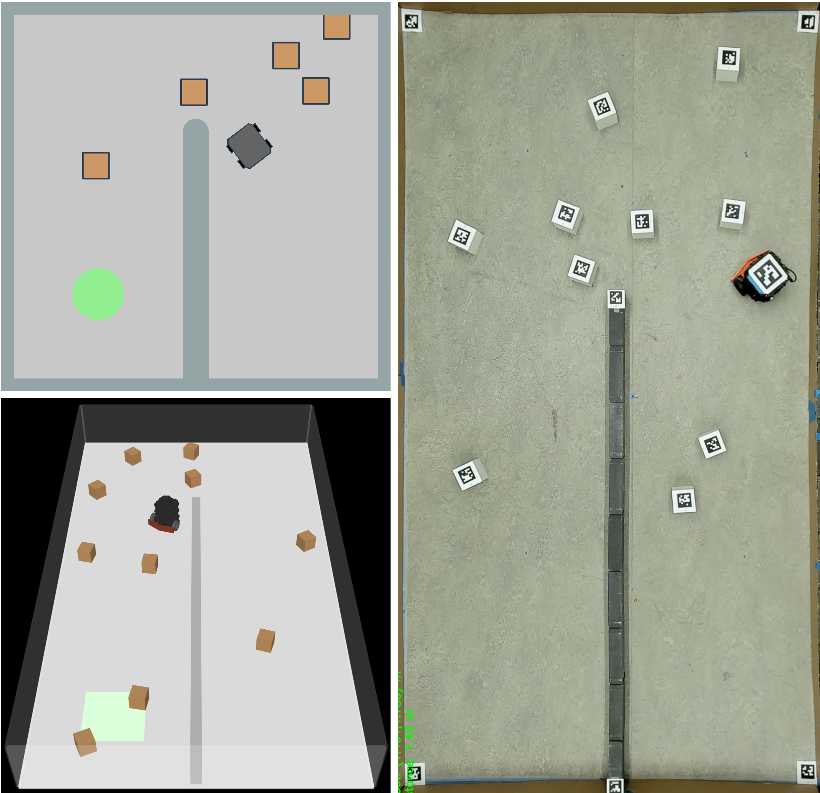}
    \caption{\textit{Maze} environment in 2D simulation (top-left panel), in 3D simulation (lower-left panel), and in our physical testbed (right panel).}
    \label{fig:2d_3d_real}
    \vspace{-5pt}
\end{figure}
\setlength{\belowcaptionskip}{1ex}

A parallel branch of work has traditionally studied such scenarios by assuming that mobile robots are equipped with manipulators and can use grasping actions to move obstacles out of the way~\cite{stilman2005navigation, petrovskaya2007probabilistic, stilman2007planning}. However, manipulators are not always viable due to cost considerations, structural limitations, or task-specific constraints. Moreover, prehensile and dexterous mobile manipulation of objects remains an area of active research. This has spurred increasing interest in pushing-based mobile robot navigation and manipulation in environments with movable objects~\cite{yao2024local, wu2020spatial}.

Despite a growing body of work on pushing-based tasks, the field still lacks a unified framework for systematic evaluation and reproducibility. Existing approaches address different aspects of the problem but share a common theme: controlling a robot to reliably push objects toward a desired direction or goal. Yet, these methods are typically tested in individually developed environments that, while similar in design, lack standardization. In addition, despite similar objectives, evaluations in existing works often rely on inconsistent metrics. As a result, potentially transferable techniques~\cite{wu2020spatial, weeda2025pushing} are only evaluated in ad hoc settings. 

Benchmarks have proven essential for advancing robotics research. Successful robotics benchmarks such as the YCB dataset~\cite{calli2015ycb}, RoboSuite~\cite{zhu2020robosuite}, and the Interactive Gibson Benchmark~\cite{xia2020interactive} have each driven significant progress in grasping, learning-based manipulation, and interactive navigation. Successful benchmarks share several guiding principles: (1) they identify the common ground across existing work; (2) they operate at an appropriate level of abstraction to ensure generalizability; and (3) they remain easy to use and configurable. Motivated by these principles and given the fragmented state of mobile robot pushing research, we present \projectname, the first comprehensive suite of tools for standardized training and evaluation of algorithms for pushing-based mobile robot tasks. \projectname~will also be made available as an open-source Python library.

\begin{figure*}[tb]
    \centering
    \includegraphics[width=\linewidth]{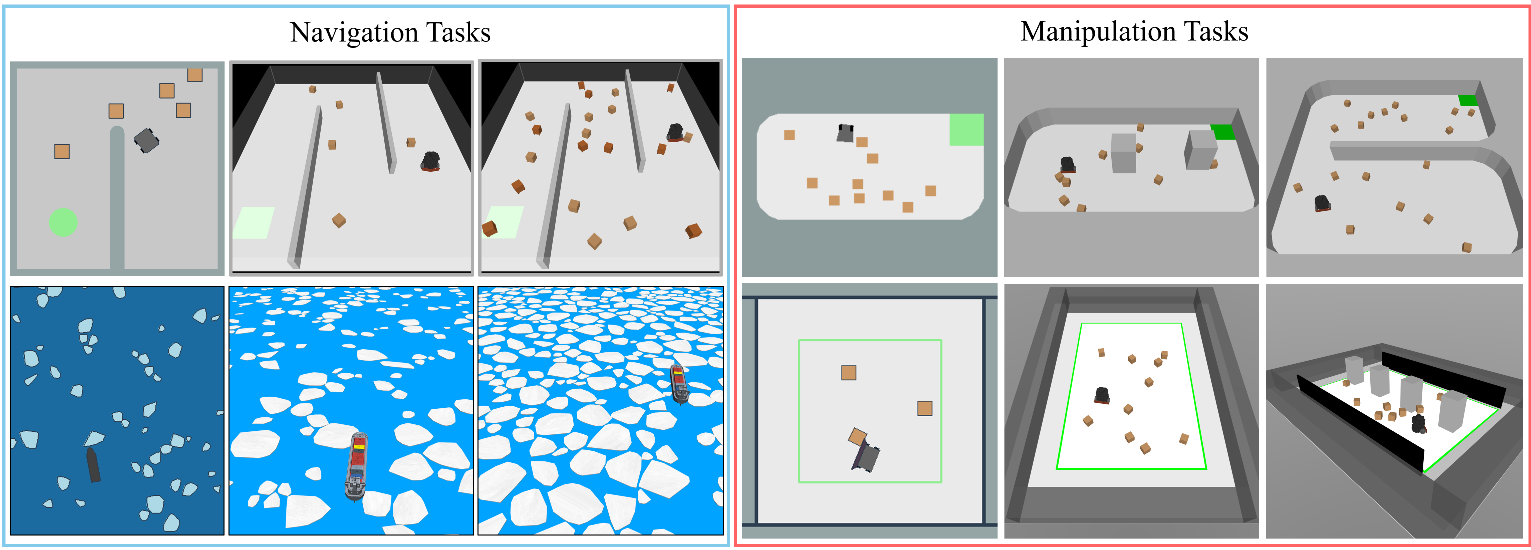}
    \caption{Illustrations of pushing-based navigation and manipulation environments in both 2D and 3D simulations, each with configurable complexity levels. \textit{Maze} (top-left): the mobile robot must reach the goal point among movable obstacles. \textit{Ship-Ice} (lower-left): an autonomous ship must navigate through ice-covered waters. \textit{Box-Delivery} (top-right): the robot needs to push all boxes into the green receptacle. \textit{Area-Clearing} (lower-right): the robot needs to remove all boxes from within a clearance area outlined by the green rectangle.}
    \label{fig:task-variations}
    \vspace{-15pt}
\end{figure*}

Our key contributions from \projectname~are as follows. First, we provide a range of simulated environments inspired by existing works in pushing-based mobile robot navigation and manipulation tasks. The environments are configurable by users to modify the difficulty. For each environment, we provide high-fidelity 3D simulations implemented using MuJoCo~\cite{todorov2012mujoco} to ultimately enable physical robot deployment, as shown in Fig.~\ref{fig:2d_3d_real}. We additionally provide 2D Pymunk~\cite{pymunk} simulations for rapid prototyping.

Second, we propose a set of novel metrics in \projectname~to evaluate and compare existing and future methods. The proposed metrics capture properties unique to mobile robot pushing-based tasks, such as efficiency and interaction trade-offs, and partial task completions.

Third, we provide reference baseline implementations in \projectname~and present their evaluation results as potential starting points for future research. Our baselines include both general reinforcement learning algorithms (SAC~\cite{haarnoja2018soft}, PPO~\cite{schulman2017proximal}) and task-specific methods such as Spatial Action Maps (SAM)~\cite{wu2020spatial} and ASV path planners~\cite{de2024auto, zhong2025autonomous}. 

Finally, we present zero-shot sim-to-real transfer with policies trained using \projectname. We show that the evaluation results from the physical robot testbed generally align with those from simulation, showcasing \projectname~as an effective workflow that scales from 2D rapid experimentation to 3D simulation and ultimately to real-world deployment.

\section{Related Work}
\label{sec:related_work}
In this section, we review related work in pushing-based mobile robot tasks and the most closely related benchmarks.

\subsection{Pushing-based Navigation and Manipulation Tasks}
Pushing-based navigation tasks focus on navigating the robot to a goal location while pushing path-blocking obstacles aside.
Recent work primarily focuses either on improving the robot's navigation adaptability in environments with pushable objects or on reducing total pushing effort. For instance, \cite{he2024interactive} leveraged online sensor feedback for adaptive pushing action selection, while~\cite{yao2024local} improved maneuverability by learning non-axis-aligned pushing. To reduce pushing effort during navigation, \cite{ren2024search} studied minimizing the number of pushes required for a mobile robot to move obstacles, while~\cite{weeda2025pushing} aims to reduce accumulated pushing force for navigation. Beyond terrestrial robots, \cite{de2024auto, zhong2025autonomous} studied autonomous surface vehicle (ASV) navigating ice-covered waters, proposing planners that reduce ice pushing.

% Recent work focuses either on efficiently maneuvering the robot through the environment while pushing objects or on minimizing the pushing effort. Navigation policies that aim to maximize efficiency typically aim to minimize a path cost or an affordance to determine whether to push or avoid an obstacle. Such a cost function can be used to design a graph-based planner~\cite{he2024interactive} or a policy using a reinforcement learning framework~\cite{yao2024local}. 

% design an updateable heuristic  in~\cite{he2024interactive}.  proposed a navigation algorithm that uses sensor feedback to update object pushing affordance for real-time action selection. The work in~\cite{} proposed a reinforcement learning method that allows for non-axis-aligned pushing for mobile robot navigating in cluttered environments with pushable objects. 

% \subsection{Pushing-based Manipulation Tasks}

Pushing-based manipulation tasks emphasize manipulating the movable objects in the environment to reach desired configurations. 
Existing work on the control level focuses on stable pushing by modeling robot-object contact. For example, \cite{tang2023unwieldy} studied maintaining a stiff robot-object contact for stable pushing, while a subsequent work~\cite{tang2024unwieldy} achieves stable pushing while the object is sliding.  Other works focus on the policy level to coordinate actions to push all objects toward some configurations. In \cite{wu2020spatial}, the Spatial Action Map (SAM) is proposed as a novel action representation suitable for multi-object pushing, where a mobile robot gathers scattered objects into a receptacle. In~\cite{xie2025autonomous}, a full-stack framework is developed for robots to autonomously push and collect trolleys in public areas.

The emergence of these diverse approaches with shared objectives underscores the need for a unified benchmark.

\subsection{Mobile Robot Benchmarks}
Traditional mobile robot benchmarks focus primarily on collision-free navigation. 
BARN~\cite{perille2020benchmarking} evaluates navigation in cluttered environments, while \cite{xu2023benchmarking} and \cite{kastner2022arena} extend beyond static environments by considering both static and dynamic obstacles. Nonetheless, these benchmarks primarily assess performance in ways that discourage environmental interaction, classifying robot-environment contact as failures rather than controlled and intended interactions.

% \cite{perille2020benchmarking} proposed BARN, a benchmark for ground mobile robot navigation in cluttered environments. Bench-MR~\cite{heiden2021bench} benchmarks sampling-based motion planning for nonholonomic mobile robots and provides diverse static environments.
%
% \cite{xu2023benchmarking} and \cite{kastner2022arena} extend beyond static environments by considering both static and dynamic obstacles. Nonetheless, these benchmarks primarily assess performance in ways that discourage environmental interaction, classifying robot-environment contact as failures rather than controlled and intended interactions.

The Interactive Gibson Benchmark~\cite{xia2020interactive} introduces interactable environments and is perhaps the closest to our work. Our benchmark differs from Gibson~\cite{xia2020interactive} in three ways. 
First, Gibson focuses \textit{only} on interactive navigation, whereas ours supports both pushing-based navigation \textit{and} manipulation tasks. 
Second, Gibson emphasizes photo-realistic rendering, which is valuable for vision research but is often heavy for rapid experimentation. In contrast, our benchmark spans both lightweight 2D simulation for pilot studies and 3D simulation for advanced evaluation. 
Third, Gibson includes only generic RL baselines, while we provide both transferable RL and task-specific baselines for stronger domain performances.
%
% Third, Gibson includes only generic RL baselines, while we provide both generic RL baselines as transferable approaches, and task-specific methods as references for stronger domain-specific performances.

Evaluation metrics in existing mobile robot benchmarks~\cite{perille2020benchmarking, xu2023benchmarking} primarily focus on navigation success rates, collision count, and path efficiency. These metrics are insufficient for pushing-based tasks where controlled interactions are required rather than avoided. Gibson \cite{xia2020interactive} introduced metrics for interactive navigation that capture both efficiency and environment disturbance from the interaction. These metrics are limited to interactive navigation and are not suitable for pushing-based manipulation where modifying the environment is part of the task (i.e., box delivery). In contrast, our benchmark introduces separate evaluation metrics for pushing-based navigation and manipulation tasks, ensuring that interaction quality and efficiency are appropriately measured in both classes.
%
% To our knowledge, ours is the first benchmark designed for pushing-based interactive tasks that encompass both navigation and manipulation.

\section{\projectname}
\label{sec:bench-NPNAMO}
\projectname~consists of three main components -- (i) environment, (ii) policy, and (iii) evaluation metrics. In this section, we detail the specifics of these environments, explain the implementation of custom policies in \projectname~and introduce the metrics used to evaluate policies. First, we clarify some of the terminology used in this section.

\textbf{Environment:} An environment is a simulated world where the robotic agent exists, integrated with the Gymnasium \cite{towers2024gymnasium} interface. Each environment is simulated in both 3D through MuJoCo~\cite{todorov2012mujoco} and 2D through Pymunk~\cite{pymunk}, and has multiple variations with increasing complexity (see Fig. \ref{fig:task-variations}). Environments can be run independently of a training session, and the robot can be manually operated by a user.

\textbf{Task:} We refer to a task as the robot's objective in a given environment. Specifically, we consider two types of tasks:
\begin{itemize}
  \item \textit{Navigation Task}, where the robot must reach a goal region while inevitably pushing path-blocking obstacles.
  \item \textit{Manipulation Task}, where the robot must push objects in the environment to some desired configurations.
\end{itemize}

\textbf{Policy:} A policy is a set of rules used to control the robot in an environment. It takes as input an observation of the environment and outputs a robot action. In \projectname, we include several baseline policies as reference implementations for users, which are listed in Table \ref{table:envs}. 

\textbf{Metric:} A metric is a score used to evaluate the effectiveness of a policy in completing a task in a given environment. In Section \ref{sec:npin-metrics}, we introduce novel metrics included in \projectname~to evaluate and compare policies.

\begin{table*}[tbp]
\caption{Interactive pushing-based tasks included in \projectname.}
\vspace{-5pt}
\label{table:envs}
\begin{center}
\setlength\tabcolsep{3pt}
\begin{tabular}{lccc}
    \toprule
     \textbf{Environments} & \textbf{Task Class} & \textbf{Variations} & \textbf{Baselines Evaluated} \\
    \midrule
    \textit{Maze} & Navigation-centric & maze complexity, obstacle count & SAC~\cite{haarnoja2018soft}, PPO~\cite{schulman2017proximal}, RRT~\cite{lavalleRandomizedKinodynamicPlanning2001}\\
    
    \textit{Ship-Ice} & Navigation-centric  &  ice concentrations & SAC~\cite{haarnoja2018soft}, PPO~\cite{schulman2017proximal}, ASV planning~\cite{de2024auto, zhong2025autonomous}\\
    
    \textit{Box-Delivery} & Manipulation-centric &  w/o static obstacles, box count & SAC~\cite{haarnoja2018soft}, PPO~\cite{schulman2017proximal}, SAM~\cite{wu2020spatial} \\

    \textit{Area-Clearing} & Manipulation-centric &  w/o static obstacles, box count & SAC~\cite{haarnoja2018soft}, PPO~\cite{schulman2017proximal}, SAM~\cite{wu2020spatial}, GTSP (see project repo)\\
    \bottomrule
\end{tabular}
\end{center}
\vspace{-15pt}
\end{table*}

\begin{table*}[tbp]
\caption{Rewards, actions, and observations for each task.}
\vspace{-5pt}
\label{table:envs_props}
\begin{center}
\setlength\tabcolsep{3pt}
\begin{tabular}{lccc}
    \toprule
     \textbf{Environments} & \textbf{Action} & \textbf{Observation} & \textbf{Rewards}\\
    \midrule
    \textit{Maze} & Angular vel.  & occupancy + footprint + goal DT & collision + distance decrement + terminal\\
    
    \textit{Ship-Ice} & Angular vel.  & occupancy + footprint + goal DT + heading encoding &  collision + heading + terminal\\
    
    \textit{Box-Delivery} & Heading & occupancy + footprint + egocentric DT + goal DT &  collision + box completion + box displacement\\

    \textit{Area-Clearing} & Heading & occupancy + footprint + egocentric DT + goal DT & collision + box completion + box displacement\\
    \bottomrule
\end{tabular}
\end{center}
\vspace{-15pt}
\end{table*}

\subsection{
\projectname~Environments}
\label{sec:npin-tasks}

We now introduce the environments and tasks in \projectname. For each environment, we provide the rationale for choosing the environment and the specifics of the setup. We also present the actions, observations, and rewards for the robotic agents, summarized in Table~\ref{table:envs_props}. Further demonstrations can also be found in the supplemental video. 

\subsubsection{Navigating maze with movable obstacles \textit{(Maze)}}
Existing work in pushing-based navigation has frequently adopted maze-like environments with bounded rectangular spaces and walls~\cite{ren2024search, weeda2025pushing, yao2024local}. These environments capture common structural layouts of many real-world indoor settings, such as offices and hospitals. Motivated by this, we introduce \textit{Maze} environment, which features a static maze structure with randomly initialized obstacles (Fig.~\ref{fig:task-variations} top-left). The mobile robot agent is a TurtleBot3 Burger, and its task is to navigate from a starting position to a goal location while minimizing path length and obstacle collisions. Variations of the environment include different maze structures, obstacle sizes, and obstacle locations and densities.

\begin{figure}
    \centering
    \includegraphics[width=0.92\linewidth]{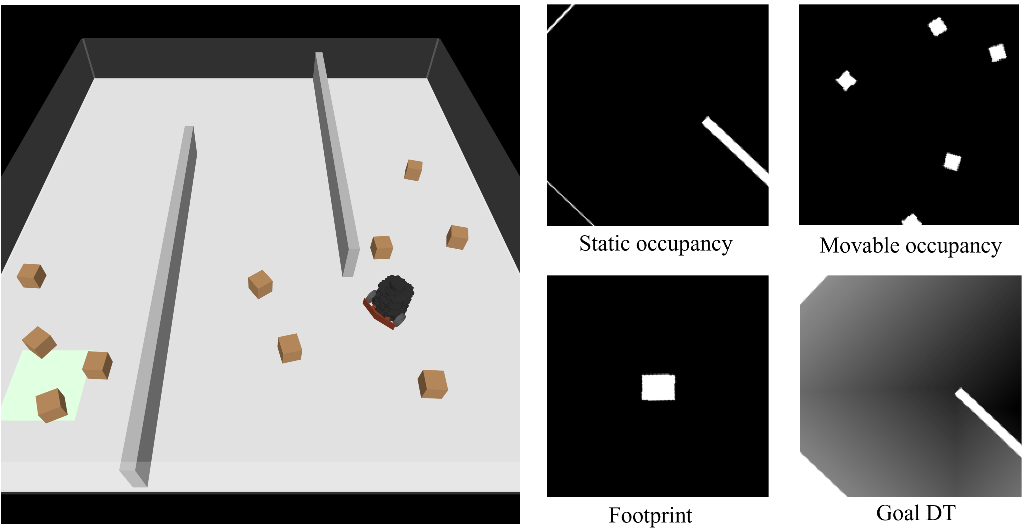}
    \caption{The \textit{Maze} environment (left) and an example egocentric observation for \textit{Maze} (right).}
    \label{fig:maze-obs}
    \vspace{-10pt}
\end{figure}

The robot in this environment is configured to move with a constant forward speed and receives an angular velocity as action input from the policy. At each timestep, the robot makes a local egocentric observation consisting of four channels that correspond to (i) static obstacle occupancy, (ii) movable obstacle occupancy, (iii) robot footprint, and (iv) a distance transform (DT) originating from the goal (goal DT). Fig.~\ref{fig:maze-obs} shows an example observation. The robot receives rewards for moving closer to the goal, penalties for colliding with obstacles, and a terminal reward when reaching the goal.

\begin{figure}
    \centering
    \includegraphics[width=0.92\linewidth]{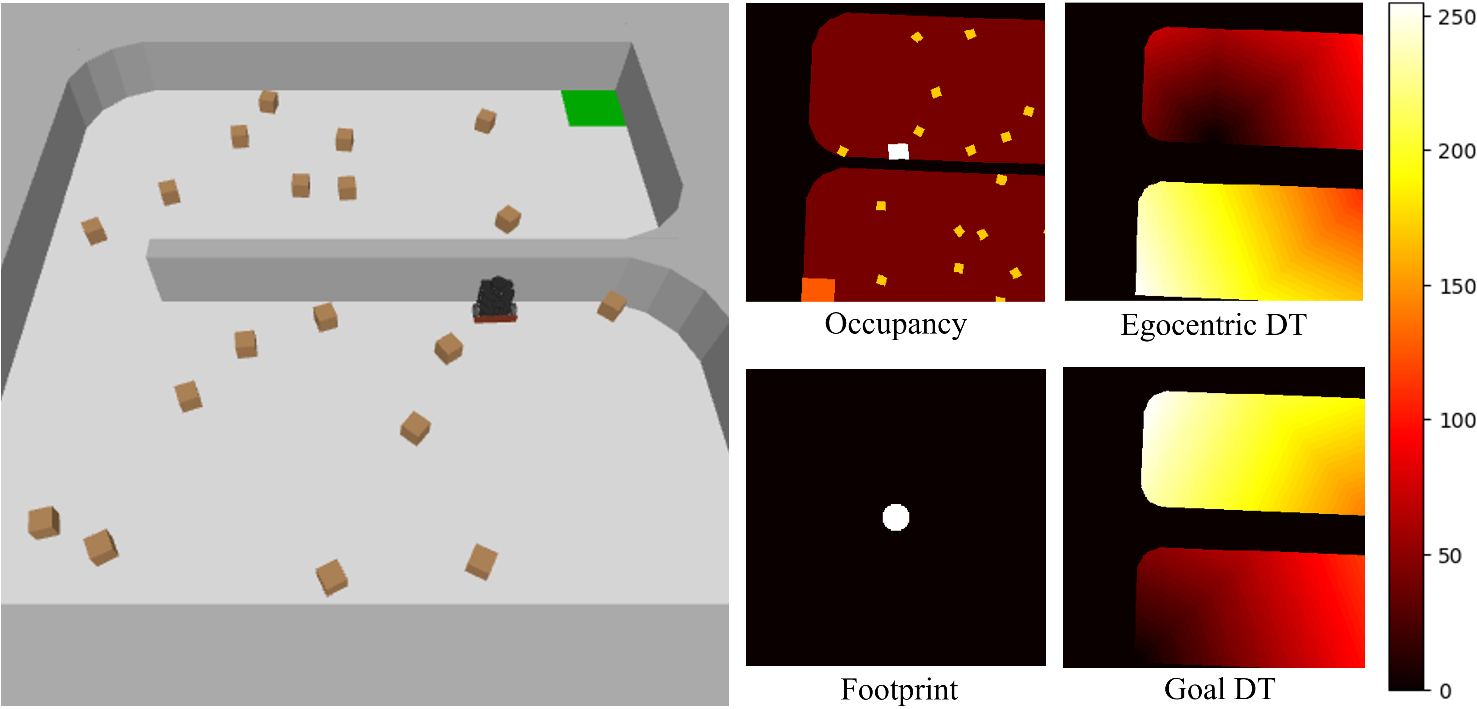}
    \caption{The \textit{Box-Delivery} environment (left) and an example observation for \textit{Box-Delivery} (right).}
    \label{fig:box-delivery-obs}
    \vspace{-15pt}
\end{figure}

\subsubsection{Autonomous ship navigation in icy waters (Ship-Ice)} 
To ensure broader applicability, we extend beyond structured indoor settings represented by \textit{Maze} to field robotics scenarios. A notable example in recent research is Autonomous Surface Vehicle (ASV) navigation through ice-covered waters~\cite{de2024auto, zhong2025autonomous}. This domain presents challenges distinct from \textit{Maze}, such as irregular obstacles, dense clustering, and fluid dynamics. For this reason, we introduce the \textit{Ship-Ice} environment, which simulates an autonomous ship navigating in a channel of ice-covered waters (Fig. \ref{fig:task-variations} bottom-left). The ship's task is to reach a horizontal goal line ahead of the ship while minimizing collisions with broken ice floes in the channel~\cite{de2024auto}. The ice floes are convex polygons with ice concentrations varying between 0\% and 50\%. 
%
% As concentrations exceed approximately 20\%, collision-free paths become non-existent, so the ship must minimize kinetic energy loss and maintain momentum. 

\textit{Ship-Ice} models ship navigation at real-world scale. We include pre-generated ice fields provided by marine study experts from [anonymous for review], while also allowing users to generate ice fields on the fly. Because MuJoCo does not natively support fluid dynamics, we approximate them on the $xy$ plane using combined linear and quadratic drag forces acting on all bodies. Relevant parameters such as ship size, mass, drag coefficients, and ice sizes are based on~\cite{de2024auto}. 

Similar to the \textit{Maze} environment, the ship moves with a constant forward speed and receives an angular velocity as action input. The ship makes observations similar to \textit{Maze} but with an additional channel that encodes the heading of the ship as a single-pixel-width line. This channel facilitates fine-grained movement around clusters of ice floes. The ship's reward function combines penalties for collisions with ice floes, a heading reward to encourage approaching the goal line, and a terminal reward for reaching the goal.

\subsubsection{Delivering boxes to a receptacle (Box-Delivery)}
Existing work in pushing-based manipulation has primarily focused on either contact-level control of pushing actions~\cite{tang2023unwieldy, tang2024unwieldy}, or task-level policies for moving all objects toward target configurations~\cite{wu2020spatial, xie2025autonomous}, with some positioned in between~\cite{bertoncelli2024streamlining}. The common ground across all these methods centers on requiring the robot to push objects in a desired direction or trajectory toward a goal. Inspired by this observation, we introduce \textit{Box-Delivery} here and \textit{Area-Clearing} in the following section (Sec.~\ref{sec:area-clearing}). The \textit{Box-Delivery} environment consists of a set of movable boxes and a designated \textit{receptacle} (Fig.~\ref{fig:task-variations} top-right). Similar to \textit{Maze}, the robotic agent is a TurtleBot3 Burger. The robot is tasked with delivering all boxes to the receptacle using a non-prehensile manipulator (e.g., front bumper) to push the boxes. The boxes and robot starting location are randomly generated within an environment. Further variations are possible by including static obstacles (e.g., columns) and changing the number and size of movable objects.

At each step, the robot receives a heading as its action input, in which case the robot travels for a fixed distance along that heading. The robot's observation is egocentric and includes (i) occupancy of the environment with static and movable objects encoded in different values, (ii) robot footprint, (iii) DT originating from the robot (egocentric DT), and (iv) DT originating from the receptacle (goal DT). 
Fig.~\ref{fig:box-delivery-obs} illustrates the environment and corresponding robot observation. The robot needs to push movable objects (brown boxes) toward the receptacle (light green square). 
The robot receives rewards for moving boxes toward the receptacle or delivering a box, and penalties when colliding with static obstacles or moving boxes away from the receptacle.

\subsubsection{Clearing boxes from an area (Area-Clearing)}
\label{sec:area-clearing}
The \textit{Area-Clearing} environment consists of a set of movable boxes and a \textit{clearance area} (Fig.~\ref{fig:task-variations} bottom-right). The robotic agent is a TurtleBot3 Burger. The task of the robot is to remove the boxes from within this clearance area, with no constraints on the final positions of the cleared boxes. Variations of this environment can be created by changing the number of objects to be removed, the number of static obstacles, and workspace constraints (e.g., walls blocking the clearance area close to the boundary).

The robot actions and observations in \textit{Area-Clearing} are the same as \textit{Box-Delivery}, except that the goal distance transform is now computed with respect to the open boundary of the clearance area.
The reward function is also similar to \textit{Box-Delivery}, where the robot receives a reward when moving any object inside the clearance area closer to the boundary, and when pushing a box out of the boundary. 

\subsection{\projectname~Implementation}
\label{sec:npin-implementation}
\projectname~is designed with a modular architecture to support flexible customization and ease of use. The benchmark has been designed with a one-line install to allow researchers to get up and running quickly. We highlight a few core aspects of \projectname’s implementation, designed to support diverse research needs and to make it convenient for users to integrate custom policies of their choice for the pushing-based tasks described above.

\subsubsection{Additional Features and Configurability}
In addition to the action spaces presented in Table~\ref{table:envs_props}, \projectname~also allows direct wheel velocity commands for the TurtleBot3 Burger in \textit{Maze}, \textit{Box-Delivery}, and \textit{Area-Clearing}, allowing for lower-level, more fine-grained control. Further, the robot can be equipped with three interchangeable front bumpers (inward-curved collector, straight-edge pusher, outward-curved navigation bumper). These bumpers potentially serve different tasks and can induce different interaction effects. Finally, in real-world indoor environments, robots often encounter both objects that rest flat on the ground (e.g., couches, tables, bags) and wheeled objects that move more easily (e.g., office chairs). To conceptually capture this challenge, the \textit{Maze}, \textit{Box-Delivery}, and \textit{Area-Clearing} environments can have both unwheeled and wheeled boxes that exhibit distinct interaction dynamics, as shown in Fig.~\ref{fig:task-variations} top-left with dark and light brown boxes.

\subsubsection{Simplified 2D Simulations}
All environments detailed in Sec.~\ref{sec:npin-tasks} are supported with a simplified simulation through Pymunk 2D~\cite{pymunk}. The 2D versions provide lightweight simulations where all bodies and interactions are modeled through convex planar polygons, making them ideal for rapid pilot studies. The 2D version of \textit{Ship-Ice} uses a model ship scale similar to the setup in~\cite{de2024auto} with no fluid dynamics. Detailed comparisons between the 2D and 3D environments can be found in the project repository.

\subsubsection{Gymnasium Integration}
We integrate all environments within the Gymnasium~\cite{towers2024gymnasium} interface to standardize policy development and evaluation. The implementation supports a wide range of environment parameters through either script-level configuration or command-line arguments. 

\subsubsection{Extensible Policy Class}
\projectname~provides a standardized, extensible policy template to simplify the incorporation of new algorithms. This template follows a plug-and-play philosophy: a user only needs to implement the required APIs within a policy class. \projectname~then handles inference, evaluation, and logging processes.

By way of illustration, \projectname~comes with reference implementations for well-established baselines, as described in Sec.~\ref{sec:experiments}. These examples demonstrate both how to integrate popular Reinforcement Learning algorithms from Stable Baselines 3~\cite{stable-baselines3} and how to incorporate specialized or state-of-the-art approaches. Additional documentation and usage examples, including configuration files and training scripts, are provided in the project repository at {\footnotesize \url{https://github.com/IvanIZ/BenchNPIN}}.

\subsection{Metrics}
\label{sec:npin-metrics}
We now provide metrics to evaluate policies for the above tasks. These metrics are inspired by~\cite{xia2020interactive} and aim to capture the task efficiency and interaction efforts of the robot. Consider a robot of mass $m_0$ that traverses a path of length $l_0$ as determined by the policy. Let $O = \{o_1, \ldots, o_K\}$ be the set of $K$ movable objects in the environment, which may have moved due to the robot's motion. Let the mass and path length for each object $i \in \{1, \ldots, K\}$ be $m_{i}$ and $l_i$, respectively. Since navigation and manipulation are intrinsically different tasks, \projectname~provides a distinct set of metrics for each class.

\subsubsection{Navigation Metrics}

\begin{figure}[t]
    \centering
    \includegraphics[width=0.78\linewidth]{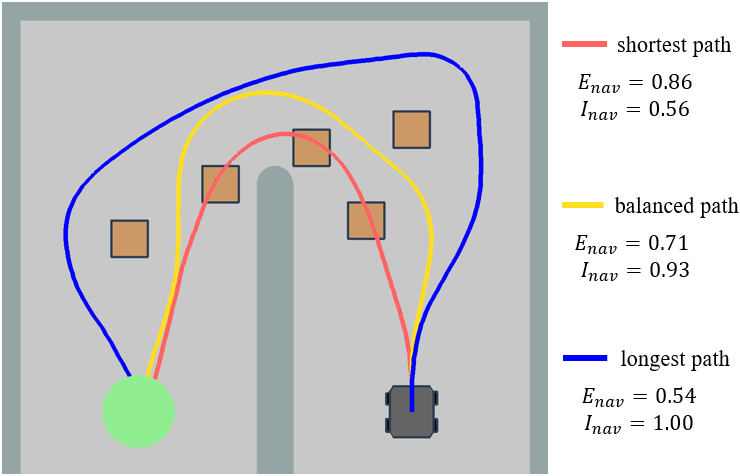}
    \caption{Three teleoperated \textit{Maze} paths and their performance. While the shortest path (red) has the highest efficiency score $E_{\text{nav}} = 0.86$, excessive collisions with the movable objects degrade the effort score $I_{\text{nav}} = 0.56$. In contrast, while the longest path (blue) is collision-free $I_{\text{nav}} = 1.00$, efficiency is largely compromised $E_{\text{nav}} = 0.54$. A balanced path (yellow) potentially offers the best trade-off.}
    \label{fig:nav-metrics-demo}
    \vspace{-15pt}
\end{figure}

We provide two metrics to evaluate navigation policies: task efficiency score $E_{\text{nav}}$ and interactive effort score $I_{\text{nav}}$. The robot's efficiency in completing a navigation task depends on the length of its path, $l_0$. Let $\mathds{1}_{\text{success}}$ be an indicator function denoting task success, which takes on a value of 1 if the robot has successfully reached the goal, and 0 otherwise. Further, let $l_0^*$ be the shortest path distance from the robot's start position to the goal that avoids only static obstacles in the environment. The efficiency score $E_{\text{nav}} \in [0, 1]$ is then computed as 
\begin{equation}
\label{eq:nav-driven-efficiency}
    E_{\text{nav}} = \mathds{1}_{\text{success}}\frac{l_0^*}{l_0}.
\end{equation}
From Eq.~\ref{eq:nav-driven-efficiency}, we see that $E_{\text{nav}} = 1$ if the robot reaches the goal with the shortest possible path length. 

The interaction effort score $I_\text{nav}$ quantifies the work done by the robot to push objects relative to the total work done. It is defined as the ratio of the work needed to move only the robot along its path to the total work expended moving both the robot and any movable objects it interacts with. Assuming a constant coefficient of kinetic friction $\mu$ across the environment, the work done to move object $i$ a distance $l_i$ is $\mu m_i g l_i$. The score $I_\text{nav} \in [0,1]$ is then the ratio of the robot's work to the total work:
\begin{equation}
\label{eq:nav-driven-effort}
    I_{\text{nav}} = \frac{\mu m_0gl_0}{\sum_{i=0}^K\mu m_igl_i} = \frac{m_0l_0}{\sum_{i=0}^Km_il_i}.
    % l_0
\end{equation} 
Here, $I_{\text{nav}} = 1$ when the robot reaches the goal without moving any objects, as the interaction terms $m_il_i$ for $i \in \{1, \ldots, K\}$ in the denominator are all zero.  $I_{\text{nav}}$ penalizes extensive interactions by the robot, such as pushing objects heavier than the robot or moving them over longer distances (increases denominator, lowers score). If the robot is significantly heavier than the objects, the work required to move the objects becomes negligible compared to the work to move the robot, which is reflected by $I_{nav}$ as it approaches 1. 

To evaluate the performance of a navigation policy, one must use both metrics to capture the trade-off between minimizing travel distance ($E_{\text{nav}}$) and avoiding unnecessary interactions ($I_{\text{nav}}$). Fig.~\ref{fig:nav-metrics-demo} illustrates how $E_{\text{nav}}$ and $I_{\text{nav}}$ capture this unique trade-off in navigation with pushing. An ideal policy will look to maximize both scores.

\begin{figure}[t]
    \centering
    \includegraphics[width=0.78\linewidth]{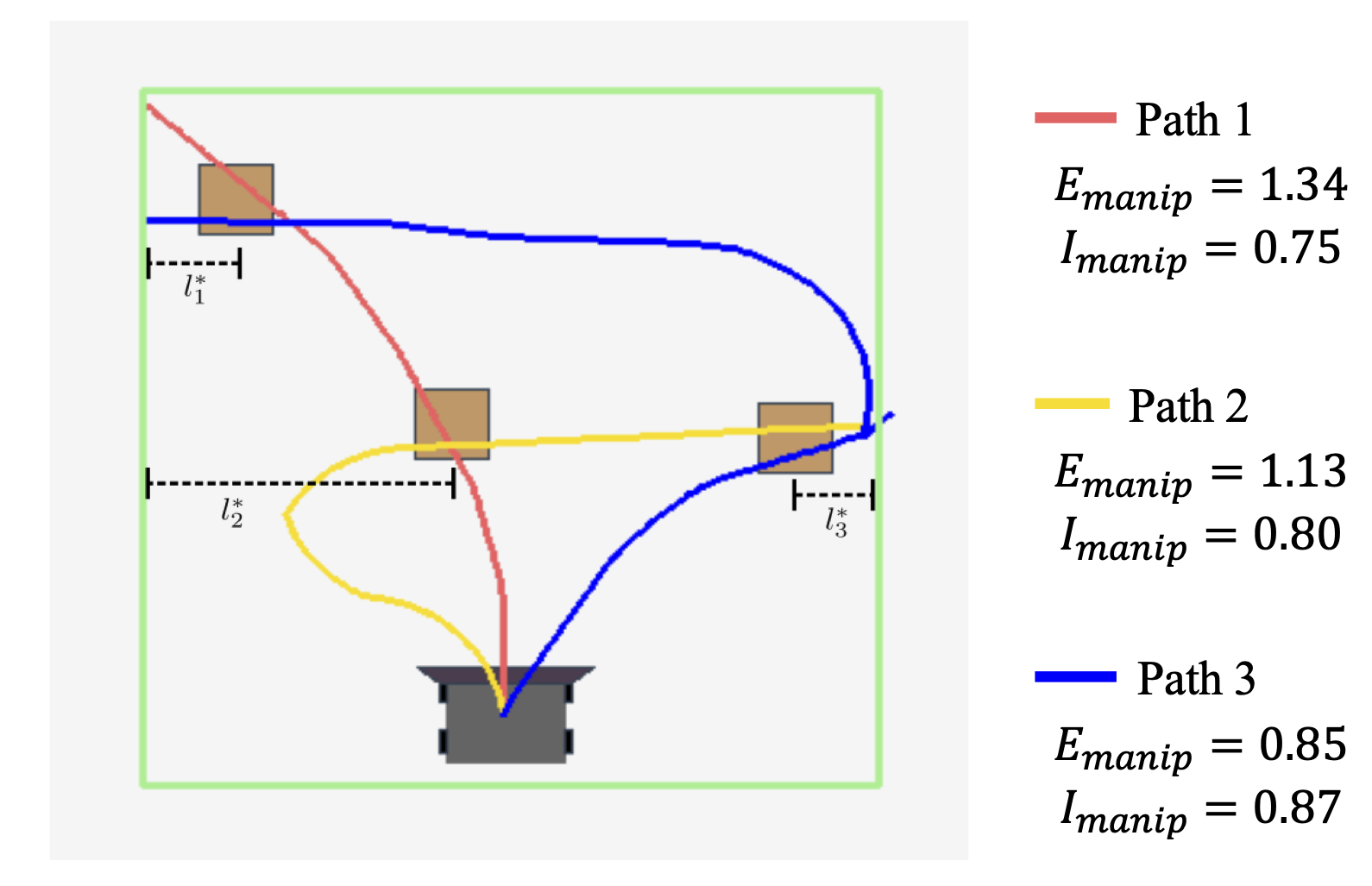}
    \caption{Three teleoperated paths for the area clearing task, where all boxes must be removed from an area (green square). Each path has a task success score $S_\text{manip} = 2/3$. While path 1 (red) has the highest efficiency score, its effort score is low as it moves a box longer than its shortest distance to the goal (dotted lines of length $l_i^*$). In contrast, path 3 (blue) achieves the best effort score by pushing as little as possible, but produces a long path and low efficiency. Path 2 (yellow) potentially offers the best trade-off.}
    \label{fig:task-metrics-demo}
    \vspace{-15pt}
\end{figure}

\subsubsection{Manipulation Metrics}
The metrics for manipulation tasks, unlike the navigation metrics, must incorporate the minimum path length and effort required by the robot to manipulate objects towards completing its task. We also look to compare robot paths that may achieve \textit{partial completion} of tasks, e.g., push two out of four boxes into the receptacle. To capture this effectively, we propose three metrics: (i) task success score $S_\text{manip}$, (ii) efficiency score $E_\text{manip}$, and (iii) interaction effort score $I_\text{manip}$.

To simplify our explanation, we consider that each of the $K$ movable objects must be manipulated to a desired set of configurations (delivered or cleared). As such, the robot must complete a series of smaller manipulation \textit{sub-tasks} corresponding to each box to complete the overall task. 
Let $\mathds{1}_{\text{success}}^i$ be an indicator function that denotes whether the sub-task corresponding to the $i^\text{th}$ movable object has been completed. Specifically, $\mathds{1}_{\text{success}}^i = 1$ if object $o_i \in O$ has been successfully pushed into the receptacle for \textit{Box-Delivery} (or removed from the clearance area for \textit{Area-Clearing}), and $\mathds{1}_{\text{success}}^i = 0$ otherwise.
Let $K' \leq K$ be the number of sub-tasks that have been completed, i.e., $K' = \sum_{i=1}^K \mathds{1}_{\text{success}}^i$. The task success score $S_\text{manip}$ is given as 
\begin{equation}
\label{eq:task-driven-success}
    S_{\text{manip}} = \frac{K'}{K}.
\end{equation}

Given that $K'$ sub-tasks were completed, let $O' \subseteq O$ be the corresponding set of completed objects, where $|O'| = K'$. Let $L^*(O')$ be (a lower bound on) the minimum path length required to complete the successful sub-tasks. For example, if the robot completes 2 out of 4 boxes, $L^*(O')$ is the minimum path length the robot must take to move the two boxes to the receptacle for \textit{Box-Delivery} (or removed from the clearance area for \textit{Area-Clearing}). The efficiency score for the manipulation task $E_\text{manip}$ is given by
\begin{equation}
\label{eq:task-driven-efficiency}
    E_{\text{manip}} = \frac{L^*(O')}{l_0}.
\end{equation}
Intuitively, $E_{\text{manip}}$ quantifies the robot's executed path length in relation to the length of an idealized shortest path to complete the $K'$ sub-tasks. In \projectname, we compute $L^*(O')$ using a minimum spanning tree (MST) of a graph $G$ that we construct as follows. The vertex set includes $K'$ vertices for the initial positions of the completed movable objects in $O'$, another $K'$ vertices for the closest point to the goal (receptacle or clearance area boundary) from each object, and a vertex for the robot. Edges are added between all pairs of objects, from the robot to each object, and from each object to its nearest goal, with the edge weights being the shortest distance between the two points. The MST of this graph is not a strict lower bound on the total path length, and as such, it is not guaranteed that $E_{\text{manip}} \in [0, 1]$. However, we find that this is a practical bound, as doubling the edges of the MST provides a path that can complete the sub-tasks (ignoring object-to-object interactions). Using both $S_{\text{manip}}$ and $E_{\text{manip}}$, we determine how many sub-tasks the robot has completed, and the robot's efficiency in completing them.

% \textcolor{blue}{Intuitively, the robot path length $l_0$ at the denominator of Eq.~\ref{eq:task-driven-efficiency} represents the absolute efficiency performance of the robot. This value is then scaled by the approximated difficulty of the completed sub-tasks captured by $L^*(O')$.} 

The interaction effort score $I_{\text{manip}}$ for manipulation quantifies the robot's effort to interact with pushable objects. Unlike in navigation, pushing objects is a necessary part of the task. So, we define $I_{\text{manip}}$ as the ratio of the minimum necessary work to complete the $K'$ sub-tasks to the total work done. We define the minimum necessary work as the work done to move the robot along its path ($\mu gm_0l_0$) plus the work required to push each successfully delivered (or cleared) object along its shortest path to the goal ($\mu g m_i l_i^*$). The total work done is the same as the denominator of $I_\text{nav}$, i.e., the work done to move the robot and push all objects. The resulting score $I_\text{manip} \in [0, 1]$ is given as
\begin{equation}
\label{eq:task-driven-effort}
    I_{\text{manip}} = \frac{m_0l_0 + \sum_{i=1}^K \mathds{1}_{\text{success}}^i m_il_i^*}{\sum^K_{i=0}m_il_i}.
\end{equation}

% For the interaction effort score, we approximate the minimum work the robot must do to complete  while traversing its path. Using the friction coefficient $\mu$ of the environment, this work is approximated as $\mu g (m_0l_0 + \sum_{i=1}^K \mathds{1}_{\text{success}}^i m_il_i^*)$, where $l_i^*$ is the shortest distance from object $i$ to the goal (see Fig.~\ref{fig:task-metrics-demo}). The actual work done by the robot is approximated as $\sum_{i=0}^K \mu m_i g l_i$, i.e., .

Similar to the navigation metrics, one must use all manipulation metrics together to evaluate a policy's ability to complete a task ($S_{\text{manip}}$), minimize path length ($E_{\text{manip}}$), and avoid unnecessary interactions ($I_{\text{manip}}$). Fig.~\ref{fig:task-metrics-demo} illustrates these metrics evaluating three trajectories for the area clearing task. All trajectories achieve a task success score of $2/3$, but with different trade-offs in path efficiency and interaction effort.
\section{Evaluating Baselines using \projectname}
\label{sec:experiments}

We now describe the baseline algorithms available through \projectname. These baselines are intended to confirm the extensibility of \projectname~and provide reference points for future studies. Further, we present physical robot experiments on a subset of tasks from Sec.~\ref{sec:npin-tasks}. Specifically, we focus on one navigation task (\textit{Maze}) and one manipulation task (\textit{Box-Delivery}), where we evaluate sim-to-real transfer with policies trained using \projectname. Due to space constraints, we include evaluation results for the remaining tasks using \projectname~simulations in the project repository. 

% \footnotetext[1]{The Appendix can be found at the project repository {\footnotesize \url{https://anonymous.4open.science/r/BenchPush-CDF6/README.md}}}

\subsection{Implemented Baselines}
\label{sec:npin-baselines}
To support both general mobile robot pushing research and task-specific studies, \projectname~includes baselines that fall into two categories: (i) reinforcement learning (RL) baselines applicable to all tasks in Section~\ref{sec:npin-tasks} and (ii) task-specific baselines. Table~\ref{table:envs} summarizes the baselines included for each task. The reinforcement learning (RL) baselines include Soft Actor-Critic (SAC)~\cite{haarnoja2018soft} and Proximal Policy Optimization (PPO)~\cite{schulman2017proximal}. These baselines were implemented using Stable Baselines 3 \cite{stable-baselines3} and trained across all tasks. To handle image observations, both baselines employ a customized ResNet18~\cite{he2016deep} backbone with the last linear layer removed as the high-dimensional observation encoder, followed by a 3-layer MLP to output actions or value estimates.

% For each task, we test the corresponding baselines using a single configuration of the environment and evaluate their performance using the metrics from Section \ref{sec:npin-metrics}.

In addition to the RL baselines, \projectname~includes specialized policies for some of the tasks, of which we highlight two policies in this paper. For \textit{Maze}, we implemented a rapidly-exploring random tree (RRT) planner based on \cite{lavalleRandomizedKinodynamicPlanning2001}, which drives the robot through the free space among pushable obstacles, albeit without considering the nature of the interactions. Secondly, for \textit{Box-Delivery} and \textit{Area-Clearing}, we adopt the Spatial Action Map (SAM) policy~\cite{wu2020spatial}, designed for multi-object pushing. The SAM policy learns to output a series of waypoints in the environment that enable the robot to complete a manipulation-centric task. To implement SAM, we configured the \textit{Box-Delivery} and \textit{Area-Clearing} environments to receive waypoints in the environment as action inputs. We additionally include task-specific baselines such as lattice-planning~\cite{de2024auto} and predictive-planning~\cite{zhong2025autonomous} policies for \textit{Ship-Ice}, and a Generalized Traveling Salesman Problem (GTSP) based policy for \textit{Area-Clearing} (details in project repository). The evaluations of these additional baselines in simulations and model weights are included in the project repository as references for future studies.

\subsection{Physical Experiment Setup}
We conducted physical robot experiments for the \textit{Maze} and \textit{Box-Delivery} as representative scenarios for navigation and manipulation tasks, respectively. The experiment testbed is shown in Fig.~\ref{fig:2d_3d_real} (right), which uses a TurtleBot3 Burger. The vision system is an overhead camera that tracks the robot and detects objects at 10\,Hz. The robot is required to interact with the 3D-printed boxes under different tasks. 

The policies deployed on the robot are trained in \projectname~3D environments with similar configurations. For both  \textit{Maze} and \textit{Box-Delivery}, we test each configuration with 20 episodes in simulation and 3 episodes in the testbed. Through the physical experiments, we seek to evaluate the effectiveness of transferring \projectname-trained policies to real hardware. Specifically, we examine whether task success rates transfer from simulation to reality, and behavioral trends on efficiency and interaction effort remain consistent. 

\subsection{\textit{Maze} Evaluations}
\label{sec:experiments-maze}
We evaluated the performance of SAC, PPO, and RRT policies in a U-shaped maze under 3-obstacle, 6-obstacle, and 10-obstacle settings, both in 3D simulation and in the physical testbed. The robot starts at one end of the maze and must reach the goal region located on the other end.

\begin{table}[tbp]
\centering
\caption{Results for \textit{Maze} in the physical testbed and \projectname~simulations. $\uparrow$ indicates higher is better.}
\label{table:maze_eval}
\setlength\tabcolsep{3pt}
\begin{tabular}{lcccc}
    \toprule
    \# Obs & Platform & Alg. & $E_{\text{nav}} \uparrow$ & $I_{\text{nav}} \uparrow$\\
    
    \midrule
    3-Obs & Sim & PPO & \textbf{0.809 $\pm$ 0.186} & 0.985 $\pm$ 0.016\\
               &         & SAC & 0.730 $\pm$ 0.307 & 0.985 $\pm$ 0.020\\
               &         & RRT & 0.779 $\pm$ 0.056 & \textbf{0.987 $\pm$ 0.010}\\
               \cmidrule(l){2-5}
               & Testbed & PPO & \textbf{0.757 $\pm$ 0.011} & \textbf{0.988 $\pm$ 0.008}\\
               &         & SAC & 0.708 $\pm$ 0.009 & 0.984 $\pm$ 0.005\\
               &         & RRT & 0.683 $\pm$ 0.001 & 0.985 $\pm$ 0.005\\

    \cmidrule(l){1-5}
    6-Obs & Sim & PPO & \textbf{0.853 $\pm$ 0.006} & \textbf{0.981 $\pm$ 0.018}\\
               &         & SAC & 0.765 $\pm$ 0.256 & 0.977 $\pm$ 0.018\\
               &         & RRT & 0.784 $\pm$ 0.051 & 0.979 $\pm$ 0.015\\
               \cmidrule(l){2-5}
               & Testbed & PPO & \textbf{0.754 $\pm$ 0.021} & 0.955 $\pm$ 0.024\\
               &         & SAC & 0.724 $\pm$ 0.023 & \textbf{0.987 $\pm$ 0.012}\\
               &         & RRT & 0.655 $\pm$ 0.005 & 0.977 $\pm$ 0.007\\

    \cmidrule(l){1-5}
    10-Obs & Sim & PPO & 0.725 $\pm$ 0.305 & 0.960 $\pm$ 0.022\\
               &         & SAC & 0.708 $\pm$ 0.298 & 0.957 $\pm$ 0.023\\
               &         & RRT & \textbf{0.758 $\pm$ 0.072} & \textbf{0.967 $\pm$ 0.018}\\
               \cmidrule(l){2-5}
               & Testbed & PPO & \textbf{0.770 $\pm$ 0.052} & \textbf{0.966 $\pm$ 0.012}\\
               &         & SAC & 0.618 $\pm$ 0.020 & 0.913 $\pm$ 0.037\\
               &         & RRT & 0.681 $\pm$ 0.022 & 0.930 $\pm$ 0.001\\
    \bottomrule
\end{tabular}
\vspace{-15pt}
\end{table}

Table~\ref{table:maze_eval} summarizes the evaluation results. Observe that results from the physical testbed overall mirror those from \projectname~simulations. In particular, interaction scores for all baselines exhibit downward trends with added obstacles, suggesting that environments with more clutter are increasingly difficult for the policies to navigate the robot. Furthermore, evaluations from both platforms indicate that PPO generally outperforms other baselines by having higher efficiency scores with comparable interaction scores, while SAC struggles the most in 10-obstacle settings.

% \textcolor{blue}{PPO and SAC also tend to outperform RRT, except for the 10 obstacle case, where RRT outperforms SAC.}

% Observe that as the number of obstacles increases, interaction effort scores decrease for PPO and SAC across all platforms, suggesting that the metric captures the fact that the algorithms struggle with increasingly cluttered environments. 

% Further, results from the physical testbed overall mirror those from \projectname~simulations: scores for both PPO and SAC exhibit similar downward trends with added obstacles, and both indicate that PPO generally outperforms SAC in efficiency and effort.
%
These observations suggest that policies trained in \projectname~transfer well to real robots, with consistent behaviors observed across simulation and physical deployment. We do note, however, that scores in the physical testbed are slightly lower than in simulation, likely reflecting sim-to-real gaps such as localization noise and control imperfections.

\subsection{\textit{Box-Delivery} Evaluations}
\label{sec:experiments-box-delivery}
We evaluated the performance of SAC, PPO, and SAM in \textit{Box-Delivery} settings, where the robot must push boxes to the receptacle with no static obstacles impeding its path. A 2D example setup is shown in the left-most \textit{Box-Delivery} snapshot in Fig.~\ref{fig:task-variations}. We performed experiments under 3-obstacle and 5-obstacle settings in both 3D simulation and in the testbed. The positions of the robot and boxes are randomly placed at the start of each trial. For the testbed experiments, a time limit of 5 minutes is enforced to ensure prompt completion of the trial.

The evaluation results are summarized in Table~\ref{table:box_delivery_eval}. We observe that SAM drastically outperforms PPO and SAC in simulation and testbed experiments, as it achieves high values for all metrics. Between the RL baselines, SAC outperforms PPO by achieving better task success and efficiency scores, with comparable interaction effort. Similar to \textit{Maze}, the metric values decrease with an increase in the number of boxes, indicating an increase in task difficulty. 

\begin{table}[tbp]
\caption{Results for \textit{Box-Delivery} in the physical testbed and \projectname~simulations. $\uparrow$ indicates higher is better.}
\label{table:box_delivery_eval}
\setlength\tabcolsep{3pt}
\begin{tabular}{lccccc}
    \toprule
    \# Obs & Platform & Alg. & $S_{\text{manip}} \uparrow$ & $E_{\text{manip}} \uparrow$ & $I_{\text{manip}} \uparrow$\\
    
    \midrule
    3-Obs &  Sim & PPO & 0.08 $\pm$ 0.18 & 0.004 $\pm$ 0.008 & \textbf{0.997 $\pm$ 0.002}\\
               &         & SAC & 0.13 $\pm$ 0.22 & 0.008 $\pm$ 0.014 & 0.985 $\pm$ 0.005\\
               &         & SAM & \textbf{1.00 $\pm$ 0.00} & \textbf{0.282 $\pm$ 0.122} & 0.987 $\pm$ 0.007\\
               \cmidrule(l){2-6}
               & Testbed & PPO & 0.00 $\pm$ 0.00 & 0.000 $\pm$ 0.000 & \textbf{0.992 $\pm$ 0.003}\\
               &         & SAC & 0.33 $\pm$ 0.27 & 0.189 $\pm$ 0.148 & 0.969 $\pm$ 0.009\\
               &         & SAM & \textbf{0.78 $\pm$ 0.16} & \textbf{0.341 $\pm$ 0.111} & 0.981 $\pm$ 0.015\\

    \cmidrule(l){1-6}
    5-Obs &  Sim & PPO & 0.20 $\pm$ 0.22 & 0.012 $\pm$ 0.013 & \textbf{0.994 $\pm$ 0.003}\\
               &         & SAC & 0.15 $\pm$ 0.17 & 0.018 $\pm$ 0.019 & 0.978 $\pm$ 0.007\\
               &         & SAM & \textbf{0.98 $\pm$ 0.06} & \textbf{0.243 $\pm$ 0.096} & 0.982 $\pm$ 0.007\\
               \cmidrule(l){2-6}
               & Testbed & PPO & 0.00 $\pm$ 0.00 & 0.000 $\pm$ 0.000 & 0.987 $\pm$ 0.003\\
               &         & SAC & 0.13 $\pm$ 0.09 & 0.093 $\pm$ 0.073 & \textbf{0.990 $\pm$ 0.001}\\
               &         & SAM & \textbf{0.67 $\pm$ 0.09} & \textbf{0.417 $\pm$ 0.089} & 0.972 $\pm$ 0.003\\
    \bottomrule
\end{tabular}
\vspace{-15pt}
\end{table}

The task success scores in the physical testbed are lower in most cases compared to the simulation, mainly due to sim-to-real gaps stemming from small controller errors that result in imperfect manipulation of the objects (e.g., boxes slipping from the robot's bumper). Despite this score reduction, the relative performances of the policies in the testbed are consistent with the simulations, indicating good policy transfer from \projectname~simulations to real robots.

\section{Conclusion, Limitations and Future Work}
\label{sec:conclusion}

We introduce \projectname, the first comprehensive benchmark for pushing-based navigation and manipulation tasks in mobile robots. \projectname~offers a suite of customizable tasks spanning both navigation and manipulation, implemented in simulation with varying levels of complexity. We also provide novel evaluation metrics that capture efficiency and interaction effort. \projectname~comes with implementations of established RL algorithms and task-specific baselines, demonstrating the benchmark’s extensibility.

% This work introduces \projectname, the first comprehensive benchmark on pushing-based navigation and manipulation tasks for mobile robots. \projectname~features various tasks in simulated environments that spans both navigation and manipulation. Users can customize tasks with different levels of complexity. Additionally, the benchmark includes implementations of both established learning policies and task-specific policies for each task, demonstrating its extensibility. It also provides a set of metrics specifically designed to measure efficiency and interaction effort within the pushing-based tasks. 

A current limitation of \projectname~is its reliance on bird's-eye environment observations and access to the knowledge of which objects are pushable. Future extensions will incorporate onboard sensing and diverse observation modalities to extend to more realistic perception pipelines.

% While these assumptions simplify the development of high-level policies, they are less realistic from a perception standpoint.

We recognize that no single benchmark can meet the needs of all researchers in mobile robot pushing. Nonetheless, given the importance of this area and the absence of standardized tools, we believe \projectname~addresses a timely need. By providing shared environments, metrics, and baselines, we hope it will serve as a foundation for advancing research on pushing-based navigation and manipulation and inspire further progress in this emerging field.

% We acknowledge that it is challenging, or even impossible, to build a single benchmark that will suit the needs of all researchers in mobile robot pushing studies. However, we have been motivated by the importance of the area and the lack of existing benchmarks. There is significant potential for progress in pushing-based tasks, and learning-based algorithms show promise.  As such, we hope that \projectname~will provide a useful tool for researchers working in mobile robot pushing, and will facilitate further research and breakthroughs in the area.
% \input{appendix}

\bibliographystyle{IEEEtran}
\bibliography{bib}

\end{document}